\documentclass[conference]{IEEEtran}

\makeatletter

\def\ps@IEEEtitlepagestyle{%
  \def\@oddfoot{\mycopyrightnotice}%
  \def\@evenfoot{}%
}
\def\mycopyrightnotice{%
  {\footnotesize XXX-X-XXXX-XXXX-X/XX/\$XX.00~\copyright~20XX IEEE\hfill}% <--- Change here
  \gdef\mycopyrightnotice{}
}

\usepackage{blindtext}
\usepackage{eso-pic}
\IEEEoverridecommandlockouts
\usepackage{cite}
\usepackage{amsmath,amssymb,amsfonts}
\usepackage{algorithmic}
\usepackage{graphicx}
\usepackage{textcomp}
\usepackage{xcolor}
\def\BibTeX{{\rm B\kern-.05em{\sc i\kern-.025em b}\kern-.08em
    T\kern-.1667em\lower.7ex\hbox{E}\kern-.125emX}}
    
\usepackage{eso-pic}
\newcommand\AtPageUpperMyright[1]{\AtPageUpperLeft{%
 \put(\LenToUnit{0.17\paperwidth},\LenToUnit{-2cm}){%
     \parbox{0.9\textwidth}{\raggedleft\fontsize{8}{11}\selectfont #1}}%
 }}%
\newcommand{\conf}[1]{%
\AddToShipoutPictureBG*{%
\AtPageUpperMyright{#1}
}
}    
\usepackage[utf8]{inputenc}
\usepackage{makecell}
\usepackage{booktabs}
\usepackage{multirow}
\usepackage{subcaption}
\usepackage{amsmath,amssymb,amsfonts}
\usepackage{algorithmic}
\usepackage{graphicx}
\usepackage{anyfontsize}                    
\usepackage{textcomp}
\usepackage{xcolor}
\usepackage{cite}
\makeatletter
\let\NAT@parse\undefined
\makeatother
\usepackage[unicode,bookmarksnumbered=true]{hyperref} %,hidelinks

\usepackage{orcidlink}
\definecolor{lightred}{RGB}{204, 255, 204}
\definecolor{lightorange}{RGB}{255,215,200}
\definecolor{lightyellow}{RGB}{153, 204, 255}
\definecolor{skyblue}{rgb}{0.53, 0.81, 0.98}
\usepackage{colortbl}
\begin{document}
\title{\vspace*{1cm}Fast 3D Foundation Model Initialized \\ Gaussian Splatting\\
\thanks{We would like to extend our sincere thanks to Aust-Agder utviklings- og kompetansefond (AAUKF) for the generous funding of the Arven etter Dannevig (The legacy of Dannevig) project, nr 62/22 which has been instrumental in the completion of this paper.}}
\author{%
\IEEEauthorblockN{1\textsuperscript{st} Anurag Dalal\,\orcidlink{0009-0007-9228-8222}}
\IEEEauthorblockA{\textit{Dep. of  Eng. Sciences}\\
\textit{University of Agder}\\
Grimstad, Norway\\
anurag.dalal@uia.no}
\and
\IEEEauthorblockN{2\textsuperscript{nd} Daniel Hagen\,\orcidlink{0000-0002-7030-6676}}
\IEEEauthorblockA{\textit{Dep. of  Eng. Sciences}\\
\textit{University of Agder}\\
Grimstad, Norway\\
daniel.hagen@uia.no}
\and
\IEEEauthorblockN{3\textsuperscript{rd} Kjell G.\ Robbersmyr\,\orcidlink{0000-0001-9578-7325}}
\IEEEauthorblockA{\textit{Dep. of Eng. Sciences}\\
\textit{University of Agder}\\
Grimstad, Norway\\
    kjell.g.robbersmyr@uia.no}
\and
\IEEEauthorblockN{4\textsuperscript{th} Kristian Muri Knausgård\,\orcidlink{0000-0003-4088-1642}}
\IEEEauthorblockA{\textit{Top Research Centre Mechatronics}\\
\textit{University of Agder}\\
Grimstad, Norway\\
kristianmk@ieee.org}
}

\maketitle
\conf{\textit{  VI. International Conference on Electrical, Computer and Energy Technologies (ICECET 2026) \\ 
6-8 July 2026, Rome-Italy}}
\thispagestyle{empty}
\pagestyle{empty}

%%%%%%%%%%%%%%%%%%%%%%%%%%%%%%%%%%%%%%%%%%%%%%%%%%%%%%%%%%%%%%%%%%%%%%%%%%%%%%%%
\begin{abstract}

This paper introduces a fast method for high-quality 3D Gaussian Splatting (3DGS) reconstruction without traditional Structure-from-Motion (SfM). The proposed approach leverages 3D Foundation Models (3DFMs) for camera pose and point-cloud initialization, then jointly optimizes both camera poses and Gaussian primitives using a depth-guided loss function. This enables fast convergence even from rough initialization with as few as 50–60 input views. To further improve reconstruction quality in sparse-view scenarios, an MLP-based pose refinement module is introduced alongside depth-guided supervision from the foundation model. Extensive experiments on Mip-NeRF 360, Tanks and Temples, and RobustNeRF demonstrate that the proposed method achieves competitive reconstruction quality (23.61 dB PSNR, 0.19 LPIPS) while reducing training time to approximately three minutes per scene. The proposed method produces ready-to-use 3DGS models at a fraction of the time required by existing pipelines, making it suitable for near real-time applications in robotics, VR, and autonomous navigation.
\end{abstract}

\section{Introduction}

3D Gaussian Splatting (3DGS)~\cite{kerbl3Dgaussians} has emerged as a powerful technique for high-quality and computationally efficient 3D scene reconstruction, transforming applications ranging from VR and robotics to autonomous navigation. Alongside, 3D Foundation Models (3DFMs) have gained tremendous popularity for geometric understanding tasks. However, most existing 3DGS pipelines rely on precise camera poses obtained from Structure-from-Motion (SfM) systems like COLMAP~\cite{schoenberger2016sfm,schoenberger2016mvs}. SfM is computationally intensive, scales quadratically with the number of input images, and can be unreliable in challenging scenarios such as low-texture scenes or limited viewpoints~\cite{lee2025dense}. This dependency not only increases overall processing time but also limits the use of 3DGS in real-time or large-scale scenarios, while propagating pose estimation errors into the reconstruction.

Recent advances in 3DFMs~\cite{wang2025vggt, dust3r_cvpr24,mast3r_eccv24} have shown promise in reducing the reliance on SfM by employing alternative methods for camera pose estimation and scene reconstruction. These methods often utilize transformer-based deep learning techniques to infer camera poses directly from multi-view images of a scene. By integrating these approaches with 3DGS, it is possible to achieve rapid initialization of the gaussian primitives in order of milliseconds and accurate 3D reconstructions without the overhead of traditional SfM pipelines, which takes a long time depending on the dataset size. VGGT-X~\cite{liu2025vggtxvggtmeetsdense}, which is based on 3DFM VGGT~\cite{wang2025vggt} is one such method that has demonstrated the ability to estimate camera poses along with a point-cloud with improved accuracy and low latency compared to COLMAP, making it a suitable candidate for integration with 3DGS.

Another technique that has shown potential in this domain is 3R-GS~\cite{huang20253rgsbestpracticeoptimizing}, which optimizes camera poses along with 3DGS. By jointly optimizing both the camera parameters and the Gaussian primitives, 3R-GS can achieve high-quality reconstructions even with a limited number of viewpoints. This approach not only reduces the dependency on accurate initial camera poses, but also enhances the overall robustness of the reconstruction process.

The proposed method combines the strengths of VGGT-X and 3R-GS to create a novel pipeline for fast 3DGS without relying on SfM. By leveraging VGGT-X for initial camera pose estimation and subsequently refining these poses using the joint optimization framework of 3R-GS, thereby achieving rapid and accurate 3D reconstructions with few viewpoints, and also being robust enough to cater for hundreds of views. This approach not only accelerates the reconstruction process but also improves the quality of the resulting 3DGS, making it suitable for near real-time applications in various fields such as virtual reality, robotics, and augmented reality.

Furthermore, to address the challenges of a sparse set of viewpoints, a depth-guided loss function is introduced that leverages geometric priors to enhance reconstruction quality. When fewer viewpoints are available, traditional photometric losses may be insufficient to constrain the optimization process, leading to ambiguous or inaccurate reconstructions. A depth-guided loss incorporates depth information from the 3DFM to provide additional geometric constraints during the joint optimization of camera poses and Gaussian parameters. This approach helps maintain geometric consistency and improves the fidelity of rendered images, particularly in regions with sparse view coverage. By combining photometric and geometric supervision, the proposed method achieves much faster reconstruction.
In summary, the contributions are as follows:

\begin{itemize}
    \item A novel pipeline that achieves fast optimization of 3DGS without the need for traditional SfM.
    \item Joint optimization of camera poses and 3DGS to enhance reconstruction quality.
    \item An unique loss function and optimization strategy tailored for low-viewpoint scenarios.
\end{itemize}

%%%%%%%%%%%%%%%%%%%%%%%%%%%%%%%%%%%%%%%%%%%%%%%%%%%%%%%%%%%%%%%%%%%%%%%%%%%%%%%%
\section{Related Work}
This section reviews the relevant literature on 3DGS and 3DFMs, focusing on their methodologies, applications, and advancements in the field.

\subsubsection{3D Gaussian Splatting}
3DGS~\cite{kerbl3Dgaussians} represents a scene as a set of anisotropic Gaussian primitives placed directly in 3D space rather than on a fixed voxel or mesh structure. Each Gaussian stores a mean (3D position), covariance (shape/orientation), color (or SH coefficients), and opacity. Rendering proceeds via forward GPU rasterization, gaussians are projected to screen space and composited front-to-back using alpha blending, enabling real-time novel view synthesis without expensive ray marching. The differentiability of the splat pipeline allows gradient-based optimization of both appearance and geometry from posed images. Compared to NeRF-style~\cite{barron2022mipnerf360} volumetric integration, 3DGS achieves large speedups in training and interactive FPS while maintaining high fidelity, thanks to adaptive spatial density and learned anisotropy that align splats with local surface structure. This efficiency makes 3DGS well suited for the accelerated pipeline where reliable poses and depth from 3DFMs seed initial Gaussian placement and reduce subsequent optimization iterations.

The efficiency of 3DGS has rapidly transformed 3D reconstruction workflows across multiple domains~\cite{10545567}. In VR/AR, its real-time rendering enables interactive scene exploration with far lower latency than NeRF-style models. In robotics and autonomous navigation, fast splat updates allow dynamic map refinement. These advances position 3DGS as an enabling geometric substrate that accelerates downstream tasks such as segmentation, tracking, and depth refinement.

\subsubsection{3D Foundation Models}
3DFMs unify multi-view perception tasks like camera pose estimation, depth prediction, point map generation, and correspondence reasoning within large transformer architectures trained on diverse large-scale image/video datasets. Most 3DFMs use large vision transformers such as DINO~\cite{caron2021emerging, oquab2023dinov2, simeoni2025dinov3} or ViT~\cite{dosovitskiy2020vit} as the backbone. This work focuses on the recent VGGT~\cite{wang2025vggt} rather than DUSt3R~\cite{dust3r_cvpr24}, or MASt3R~\cite{mast3r_eccv24} on accelerated 3DGS pipelines because of its faster inference time and better accuracy.

DUSt3R~\cite{dust3r_cvpr24} focuses on dense multi-view matching and geometry reconstruction through learned pixel-aligned descriptors and global alignment. It produces relative camera poses and a fused point-cloud by aggregating per-view features into a shared 3D latent representation. 
MASt3R~\cite{mast3r_eccv24} extends multi-view reconstruction with improved attention aggregation over views, yielding refined depth, normal, and point map predictions along with camera geometry. Its multi-view fusion improves completeness and reduces holes relative to single-view depth estimators. For downstream tasks: normal and depth enhance open-world segmentation by improving region separation (geometry-aware prompting~\cite{ai2025gaprompt}), consistent multi-view depth can bootstrap indoor/outdoor scale estimation modules, and predicted surface orientation can modulate splat anisotropy by aligning Gaussian covariance with local tangent planes for sharper renderings with fewer splats.

\begin{figure*}[tb!]
    \centering
    \includegraphics[width=2\columnwidth]{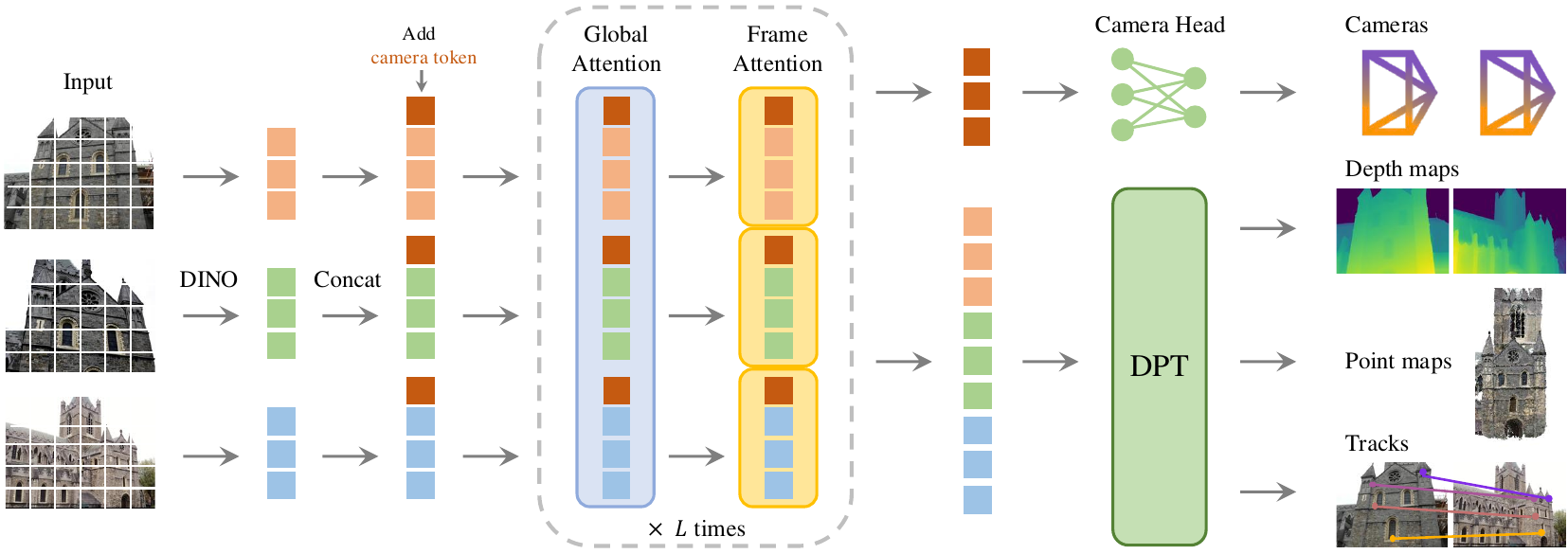}
    \caption{VGGT.~\cite{wang2025vggt} architecture illustrating multi-view fusion and task heads.}
    \label{fig:architecture_vggt_v4}
\end{figure*}

VGGT~\cite{wang2025vggt} proposes a unified transformer with DINO as its backbone, as shown in Fig.~\ref{fig:architecture_vggt_v4} that ingests a set of unordered multi-view images and jointly infers (a) camera extrinsics/intrinsics, (b) per-pixel depth maps, (c) dense tracking features, and (d) point maps or normalized 3D coordinates. It leverages cross-frame attention to build a geometry-aware latent space without an explicit SfM step. The model learns implicit epipolar constraints through large-scale pretraining and can output camera parameters in a single forward pass. VGGT-X~\cite{liu2025vggtxvggtmeetsdense} further improves pose accuracy and depth quality by integrating dense feature matching and geometric consistency losses during fine-tuning. Its enhanced pose estimates and depth maps provide reliable initialization for downstream 3D reconstruction tasks, making it well suited for the SfM-free 3DGS pipeline.

\begin{figure*}[tb!]
    \centering
    \includegraphics[width=2\columnwidth]{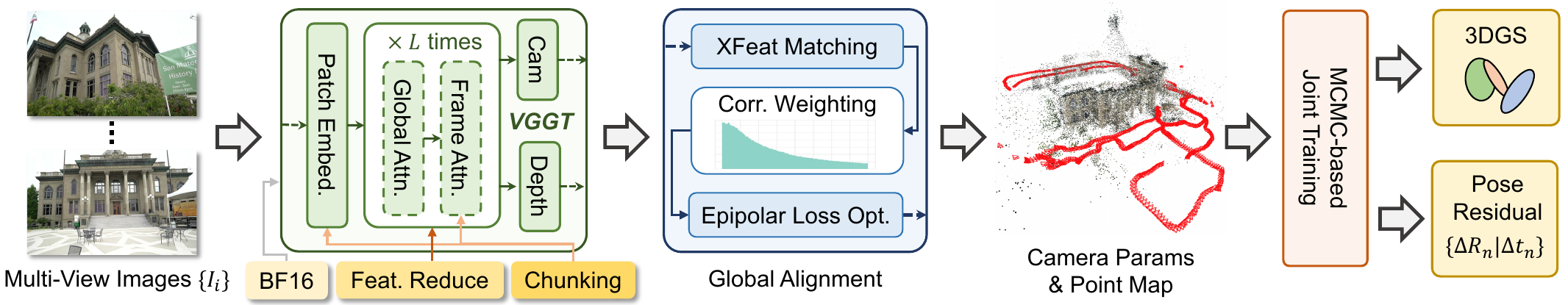}
    \caption{The pipeline diagram of VGGT-X~\cite{liu2025vggtxvggtmeetsdense}, that refines over VGGT using MCMC based joint training.}
    \label{fig:vggtx_architecture}
\end{figure*}
In summary, 3DFMs provide a unified set of geometric primitives like poses, depth, normals, point maps, and confidence that dramatically compress the traditionally sequential SfM+MVS pipeline into one forward pass, while enabling richer downstream adaptation for depth refinement, semantic segmentation, and for this case accelerated Gaussian Splatting.

\section{Methodology}
This section details the proposed pipeline that integrates VGGT-X and 3R-GS for fast  3DGS without relying on SfM. Finally, the unique loss function and training strategy that is utilized to handle fewer viewpoints scene as well aids in achieving faster convergence is discussed.

\subsection{Proposed Pipeline}

\subsubsection{Camera Parameter and Depth Estimation Mechanism} The original 3DGS~\cite{kerbl3Dgaussians} is highly sensitive to initialization. Because Gaussian primitives have limited ability to move far from their starting positions, small pose or point-cloud errors can trap them in poor local minima. Photometric losses provide only local gradients, making it difficult for misaligned Gaussians to correct themselves. Additionally, the 3DGS densification strategy is based on hand-tuned gradient thresholds, which becomes unstable when also optimizing camera poses. VGGT processes a batch of images through (1) patch embedding, (2) multi-view cross-attention blocks, and (3) task heads. Cross-attention mixes tokens across frames, allowing the model to infer relative orientation and translation by optimizing consistency of learned geometric tokens. Pose head outputs extrinsics (rotation, translation) and intrinsics or focal length. The depth head predicts the depth per-pixel aligned to those poses; a point map head may output scene coordinates in a canonical frame. Internally, learned attention patterns approximate epipolar reasoning: correspondences produce constraints the transformer aligns implicitly, akin to neural bundle adjustment without iterative classical optimization. Confidence maps or uncertainty estimates further filter unreliable regions, enabling robust initialization for downstream optimization.

Building on top of VGGT, VGGT-X~\cite{liu2025vggtxvggtmeetsdense}, which incorporates a memory-efficient VGGT implementation that scales to 1,000+ images and an adaptive global alignment for VGGT output enhancement, is used for camera pose initialization and getting the depth map. The 
architecture is shown in Fig.~\ref{fig:vggtx_architecture}
\begin{table}[tb!]
	\centering
	\caption{Learning rates for Gaussian parameters}
    \footnotesize
	\begin{tabular}{l c}
		\toprule
		Parameter & Learning Rate \\
		\midrule
		Mean & 1.6e-4 times scene scale \\
		Log-scales & 5e-3 \\
		Quaternion rotation & 1e-3 \\
		Opacity logit & 5e-2 \\
		SH color coefficients (DC) & 2.5e-3 \\
		SH color coefficients (higher orders) & 1.25e-4 \\
		\bottomrule
	\end{tabular}
	\label{tab:lr}
\end{table}
\begin{table*}[t]
    \centering
    \caption{Baseline training metrics for training and validation set. Metrics with $\uparrow$ are higher-better; $\downarrow$ lower-better.}
    \begin{tabular}{l r r r r r r r r r r}
		\toprule
        Scene & \makecell{Time\\(s)} & \makecell{Mem\\(GB)} & \makecell{Train\\PSNR $\uparrow$} & \makecell{Train\\SSIM $\uparrow$} & \makecell{Train\\LPIPS $\downarrow$} & \makecell{Val\\PSNR $\uparrow$} & \makecell{Val\\SSIM $\uparrow$} & \makecell{Val\\LPIPS $\downarrow$} & \makecell{ATE $\downarrow$} & \makecell{RTE $\downarrow$} \\
        \midrule
        bicycle   & 154.18 & 9.21 & 22.26 & 0.7103 & 0.2003 & 24.22 & 0.7103 & 0.2003 & 0.0048 & 0.4078 \\
        drjohnson & 143.16 & 9.92 & 23.63 & 0.5666 & 0.4842 & 19.69 & 0.5666 & 0.4842 & 0.0952 & 4.8973 \\
        stump     & 141.77 & 9.92 & 24.53 & 0.4966 & 0.2696 & 22.59 & 0.4966 & 0.2696 & 0.0117 & 0.3843 \\
        bonsai    & 173.54 & 9.58 & 26.89 & 0.7979 & 0.1266 & 25.99 & 0.7979 & 0.1266 & 0.0052 & 0.7599 \\
        garden    & 151.84 & 9.58 & 26.75 & 0.6122 & 0.1415 & 22.65 & 0.6122 & 0.1415 & 0.0016 & 0.3844 \\
        room      & 159.79 & 9.92 & 29.39 & 0.8921 & 0.0897 & 28.48 & 0.8921 & 0.0897 & 0.0073 & 0.6652 \\
        android   & 184.26 & 9.92 & 24.26 & 0.7977 & 0.1442 & 23.50 & 0.7977 & 0.1442 & 0.0093 & 0.5867 \\
        counter   & 167.78 & 9.58 & 26.80 & 0.8293 & 0.1097 & 26.16 & 0.8293 & 0.1097 & 0.0046 & 0.6045 \\
        crab1     & 178.50 & 9.92 & 25.65 & 0.8581 & 0.1725 & 23.80 & 0.8581 & 0.1725 & 0.0109 & 1.7140 \\
        playroom  & 140.36 & 9.92 & 25.95 & 0.7936 & 0.2020 & 25.04 & 0.7936 & 0.2020 & 0.0531 & 2.2704 \\
        kitchen   & 175.10 & 9.58 & 26.41 & 0.7485 & 0.1180 & 24.52 & 0.7485 & 0.1180 & 0.0052 & 0.4322 \\
        truck     & 166.74 & 9.92 & 23.57 & 0.8057 & 0.1275 & 22.28 & 0.8057 & 0.1275 & 0.0049 & 0.4924 \\
        crab2     & 249.77 & 16.48 & 24.97 & 0.8387 & 0.2045 & 24.12 & 0.8387 & 0.2045 & 0.0147 & 2.4636 \\
        yoda      & 281.29 & 16.72 & 26.15 & 0.8423 & 0.2197 & 25.33 & 0.8423 & 0.2197 & 0.0155 & 2.6221 \\
        statue    & 226.84 & 16.48 & 20.39 & 0.7374 & 0.2265 & 20.09 & 0.7374 & 0.2265 & 0.0077 & 0.2853 \\
        train     & 184.20 & 9.92 & 21.24 & 0.6839 & 0.2347 & 19.24 & 0.6839 & 0.2347 & 0.0057 & 0.5182 \\
        \midrule
        AVG       & 179.94 & 11.04 & 24.93 & 0.8228 & 0.1683 & 23.61 & 0.7507 & 0.1919 & 0.0161 & 1.2180 \\
        \bottomrule 
    \end{tabular}
	\label{tab:baseline_mcmc}
\end{table*}
\begin{table*}
    \centering
    \caption{Quantitative comparison of novel view synthesis. (-) denotes unreported results for ZeroGS.}
    \footnotesize
\begin{tabular}{cccclccclccclccc}
\toprule
                         & \multicolumn{3}{c}{3DGS} &  & \multicolumn{3}{c}{ZeroGS} &  & \multicolumn{3}{c}{3RGS} &  & \multicolumn{3}{c}{Ours} \\ 
\cmidrule{2-4} \cmidrule{6-8} \cmidrule{10-12} \cmidrule{14-16}
\multirow{-2}{*}{Scenes} & PSNR$\uparrow$ & SSIM$\uparrow$ & LPIPS$\downarrow$ &  & PSNR$\uparrow$ & SSIM$\uparrow$ & LPIPS$\downarrow$ &  & PSNR$\uparrow$ & SSIM$\uparrow$ & LPIPS$\downarrow$ &  & PSNR$\uparrow$ & SSIM$\uparrow$ & LPIPS$\downarrow$ \\ 
\midrule
garden  & 24.85 & 0.729 & 0.126 &  & 25.47 & 0.839 & 0.107 &  & 26.44 & 0.820 & 0.131 &  & 22.65 & 0.6122 & 0.1415 \\
counter & 27.57 & 0.862 & 0.209 &  & 26.87 & 0.873 & 0.124 &  & 28.80 & 0.897 & 0.157 &  & 26.16 & 0.8293 & 0.1097 \\
bicycle & 17.52 & 0.303 & 0.567 &  & 23.10 & 0.707 & 0.201 &  & 24.89 & 0.727 & 0.252 &  & 24.22 & 0.7103 & 0.2003 \\
room    & 30.66 & 0.899 & 0.204 &  & -     & -     & -     &  & 31.82 & 0.924 & 0.154 &  & 28.48 & 0.8921 & 0.0897 \\ 
\bottomrule
\end{tabular}
    \label{tab:nvs}
\end{table*}

\begin{table*}
    \centering
    \caption{Quantitative comparison of camera pose registration. (-) indicates unreported results.}    
    \label{tab:pose}
    \footnotesize
\begin{tabular}{cccccccccccccc}
\toprule
Scenes & \multicolumn{2}{c}{3DGS} &  & \multicolumn{2}{c}{ZeroGS} &  & \multicolumn{2}{c}{3RGS} &  & \multicolumn{2}{c}{Ours} \\ 
\cmidrule{2-3} \cmidrule{5-6} \cmidrule{8-9} \cmidrule{11-12}
       & Rotation(°)$\downarrow$ & ATE(m)$\downarrow$ &  & Rotation(°)$\downarrow$ & ATE(m)$\downarrow$ &  & Rotation(°)$\downarrow$ & ATE(m)$\downarrow$ &  & Rotation(°)$\downarrow$ & ATE(m)$\downarrow$ \\ 
\midrule
garden  & 0.19 & 0.003 &  & 0.03 & 0.002 &  & 0.03 & 0.002 &  & 0.3844 & 0.0016 \\
counter & 0.25 & 0.011 &  & 0.03 & 0.002 &  & 0.05 & 0.003 &  & 0.6045 & 0.0046 \\
bicycle & 1.07 & 0.034 &  & 0.04 & 0.005 &  & 0.09 & 0.013 &  & 0.4078 & 0.0048 \\
room    & 0.27 & 0.016 &  & -    & -     &  & 0.13 & 0.012 &  & 0.6652 & 0.0073 \\ 
\bottomrule
\end{tabular}
\end{table*}
\subsubsection{Camera Pose Optimization} Gaussian primitives have a limited ability to move far from their initial positions. When the starting point-cloud or camera poses contain errors, the photometric rendering loss provides only local gradients, which makes it difficult for a misaligned Gaussian to escape poor local minima. This often results in sub-optimal convergence and degraded scene reconstruction quality. Furthermore, adaptive density control in 3DGS relies on gradient-magnitude thresholds that require manual tuning, which becomes increasingly unstable when jointly optimizing Gaussian primitives and camera poses.

To address these issues, the 3DGS-MCMC strategy is employed~\cite{kheradmand20253dgaussiansplattingmarkov}, which reformulates Gaussian optimization as Markov Chain Monte Carlo sampling. The training process is interpreted as sampling from a distribution \( p(G) \) that assigns higher probability to Gaussian configurations that accurately reproduce the training images. Under this formulation, standard 3DGS optimization behaves similarly to Stochastic Gradient Langevin Dynamics (SGLD), with updates of the form:
\begin{equation}
G \leftarrow G + a \,\nabla_G \log p(G) + b\,\eta,
\label{eq:mcmc_update}
\end{equation}
where \( \eta \) is exploration noise and \( a \) and \( b \) control the balance between convergence and exploration. The injected noise enables Gaussians to escape local minima caused by imperfect initialization. Moreover, 3DGS-MCMC replaces heuristic densification and pruning strategies with principled state transitions, and incorporates a regularizer that encourages parsimonious Gaussian usage. Together, these components allow robust joint optimization of camera poses and Gaussian primitives.

\begin{table*}
    \centering
    \caption{Ablation study comparing training/validation reconstruction metrics and pose errors.}
    \footnotesize
    \label{tab:our_ablation}
\begin{tabular}{lcccccccc}
\toprule
Configuration & Train PSNR$\uparrow$ & Train SSIM$\uparrow$ & Train LPIPS$\downarrow$ 
      & Val PSNR$\uparrow$ & Val SSIM$\uparrow$ & Val LPIPS$\downarrow$
      & ATE$\downarrow$ & RTE$\downarrow$ \\
\midrule
Full Method   
& 24.93 & 0.8228 & 0.1683 
& 23.61 & 0.7507 & 0.1919 
& 0.0161 & 1.2180 \\
w/o MCMC 
& 21.58 & 0.7242 & 0.3030
& 20.99 & 0.6796 & 0.3063
& 0.0156 & 1.1560 \\
w/o Depth  
& 25.87 & 0.8316 & 0.1679
& 21.58 & 0.6711 & 0.2584
& 0.0211 & 1.2549 \\
\bottomrule
\end{tabular}
\end{table*}

\begin{figure}
    \centering
    \includegraphics[width=1\linewidth]{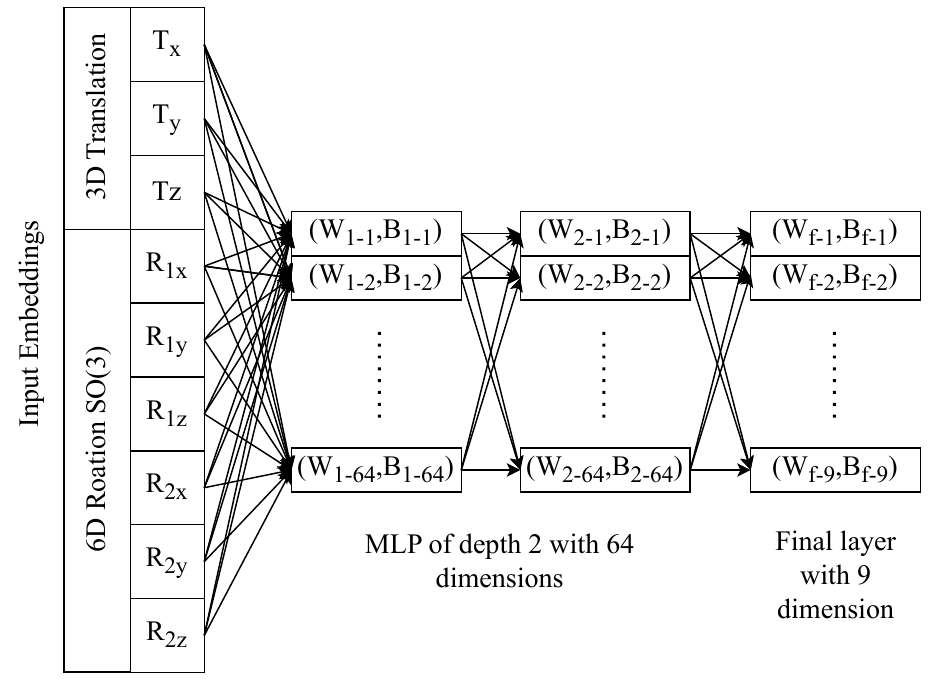}
    \caption{The Pose MLP structure that is used for every camera input of the scene.}
    \label{fig:mlp}
\end{figure}

Despite improved robustness, camera poses may still exhibit shared global drift: even if relative poses are accurate, the entire camera set may be shifted or rotated away from the true configuration. Directly optimizing each pose independently ignores these global correlations and can distort otherwise correct relative geometries due to the inherent non-convexity of pose optimization. To address this, an MLP-based global pose refiner is proposed similar to 3R-GS~\cite{huang20253rgsbestpracticeoptimizing} that predicts pose corrections from a learned camera embedding. For each camera \( i \), the correction is given by:
\begin{equation}
\Delta T_i = R_{\text{MLP}}(z_i),
\label{eq:pose_refinement}
\end{equation}
where \( z_i \) is a learnable embedding associated with camera \( i \), and the output consists of translation \( \Delta t_i \in \mathbb{R}^3 \) and rotation \( \Delta r_i \in \mathbb{R}^6 \) components. The MLP depicted in Fig.~\ref{fig:mlp}, is initialized with a zero-mean prior to ensure stable and unbiased refinement. By sharing the refinement network across all cameras, the method captures global pose relationships and improves the accuracy of camera pose adjustment, outperforming direct per-camera optimization. 

\subsubsection{Loss Function} 3DGS uses a combination of loss $\mathcal{L}_1$ and a Structural Similarity Index Measure (SSIM) loss~\cite{wang2004image}, to supervise rendered images against ground truth. The photometric loss is defined as: 
	\begin{equation}
		\mathcal{L}_{\text{Photo}} = (1 - \lambda_{\text{SSIM}}) \mathcal{L}_{\text{L1}} + \lambda_{\text{SSIM}} \mathcal{L}_{\text{SSIM}},
	\end{equation}  
where $\mathcal{L}_{\text{L1}}$ is the Mean Absolute Error (MAE) between the rendered image $I_{\text{render}}$ and the ground truth image $I_{\text{GT}}$:
		\begin{equation}
			\mathcal{L}_{\text{L1}} = \frac{1}{N} \sum_{i} \left| I_{\text{render}}(i) - I_{\text{GT}}(i) \right|,
		\end{equation}
and $\mathcal{L}_{\text{SSIM}}$ is the SSIM loss between $I_{\text{render}}$ and $I_{\text{GT}}$.

A depth guided loss term ($\mathcal{L}_{\text{Depth}}$) is also introduced along with $\mathcal{L}_{\text{Photo}}$. In 3DGS, the alpha channel can be used to encode accumulated visibility or effective depth ordering information during the splat compositing process. Unlike traditional rasterization that relies on a hardware depth buffer, Gaussian splatting renders semi-transparent 3D Gaussians in a front to back order using alpha blending. Here, alpha represents the opacity contribution of each Gaussian, which implicitly encodes how much of the background remains visible. As splats accumulate, the alpha channel stores the running transmittance, effectively capturing depth dependent occlusion information. In this sense, alpha acts as a soft, differentiable substitute for a hard Z-buffer, enabling smooth blending, correct visibility, and gradient-based optimization during training. This loss uses the normalized alpha channel of the Gaussian Splatting render ($D_{\text{render}}$) and the normalized depth maps from VGGT-X ($D_{\text{VGGTX}}$):
\begin{equation}
	\mathcal{L}_{\text{Depth}} = \frac{1}{N} \sum_{i} \left| D_{\text{render}}(i) -  D_{\text{VGGTX}}(i) \right|.
\end{equation}
The total loss is defined as:  
\begin{equation}
	\mathcal{L}_{\text{Total}} = \mathcal{L}_{\text{Photo}} + \lambda_{\text{Depth}} \mathcal{L}_{\text{Depth}}.
\end{equation}  

\subsection{Integration} 
\begin{figure*}[tb!]
    \centering
    \includegraphics[width=1.8\columnwidth]{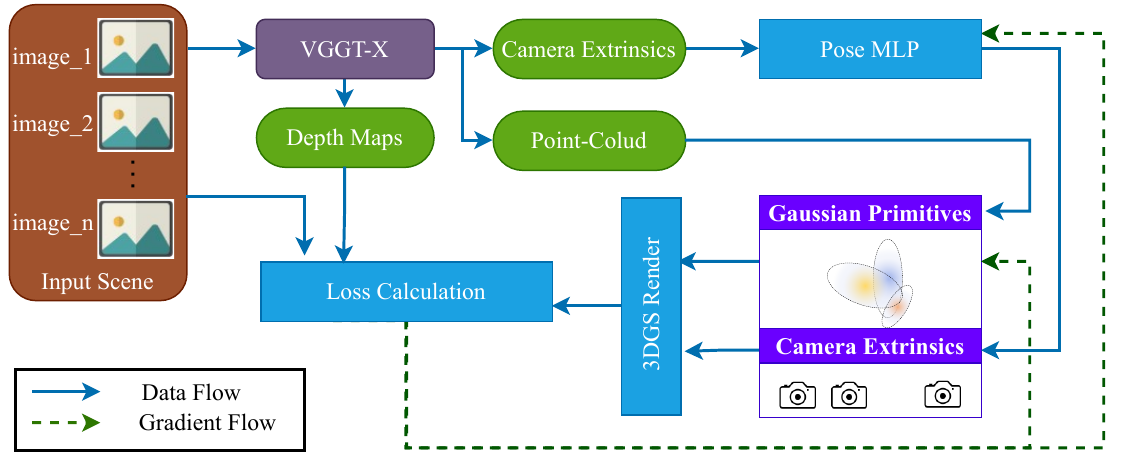}
    % \includesvg[width=2\columnwidth]{figures/architecture.svg}
    \caption{The proposed pipeline: VGGT-X estimates camera poses and depth maps from multi-view images, which initialize the 3DGS model for joint optimization of Gaussian primitives and camera poses using a depth-guided loss function.}
    \label{fig:architecture}
\end{figure*}

First, the multi-view images are passed through VGGT-X to obtain initial camera parameters and depth. These camera poses and depths serve as high-quality initializations for 3D Gaussian Splatting by seeding Gaussian centers from back-projected depth, constraining early optimization with reliable intrinsics and extrinsics, and pruning outlier pixels using confidence masks produced by the transformer. Fig.~\ref{fig:architecture} illustrates the overall pipeline.
\section{Experiments}
This section presents a comprehensive evaluation of the proposed method.\footnote{Code available at: \url{https://github.com/anurag-dalal/FastSplatting}} First, the datasets used for the evaluation are described. The training strategies and implementation details are then outlined. Finally, quantitative and qualitative results are presented, followed by ablation studies.
\begin{figure*}[htb!]
    \centering
    \includegraphics[width=2\columnwidth]{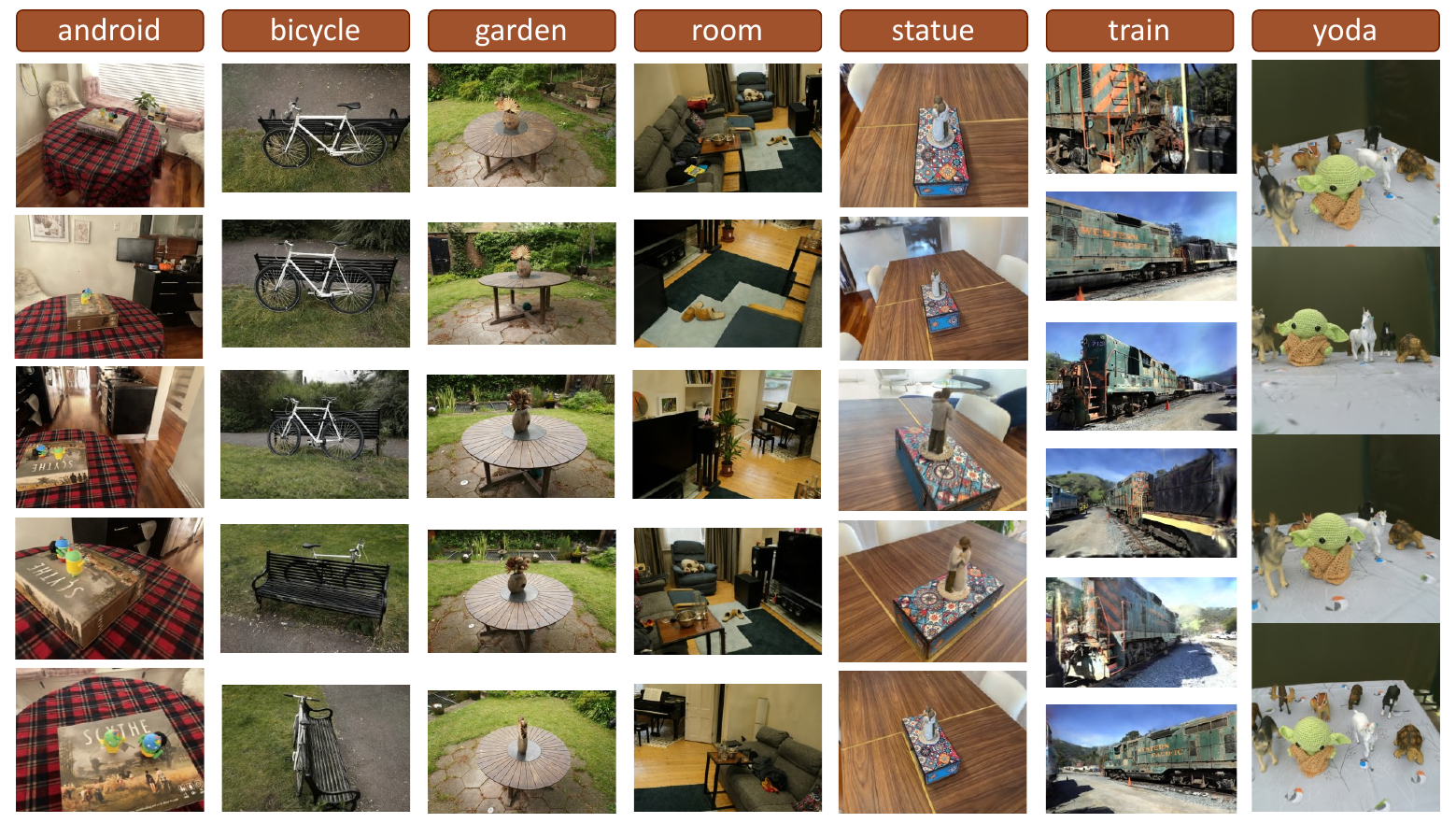}
    \caption{Rendering results from the proposed method on various scenes in the selected datasets.}
    \label{fig:rendering}
\end{figure*}
\subsection{Datasets}
The proposed method is evaluated on several publicly available datasets that are commonly used for 3D reconstruction and novel view synthesis tasks. These datasets offer a variety of scenes, viewpoints, and complexities to comprehensively assess the performance of the method.

\subsubsection{Mip-NeRF 360} This dataset~\cite{barron2022mipnerf360} consists of real-world scenes captured with a camera performing a full 360-degree rotation around a central point of interest. Each scene features complex geometry and rich appearance details, including reflective surfaces, foliage, and cluttered backgrounds that challenge view synthesis methods. The dataset provides high-resolution images with consistent exposure settings to minimize photometric variation during capture. By spanning both indoor and outdoor environments, Mip-NeRF 360 offers diverse lighting conditions, occlusions, and depth ranges. Its full-360 capture setup and challenging visual complexity make it a widely used benchmark for evaluating 3D reconstruction, neural radiance field methods, and Gaussian splatting approaches.

\subsubsection{Tanks and Temples} This dataset~\cite{Knapitsch2017} is a widely adopted benchmark to evaluate real-world 3D reconstruction and novel view synthesis methods. In these experiments, scenes from the supplementary database subset are used, denoted as \texttt{tandt\_db}. The \texttt{tandt\_db} subset includes challenging indoor environments such as \texttt{drjohnson}, which features museum-like sculptures and complex lighting, and \texttt{playroom}, a cluttered indoor room with diverse objects and occlusions. From the main T\&T benchmark, the \texttt{train} scene and the outdoor \texttt{truck} scene is included, the latter being notable for its large-scale geometry, vegetation, and detailed background structures. Together, these scenes provide a diverse set of indoor and outdoor conditions, enabling a rigorous evaluation of reconstruction robustness and generalization.

\subsubsection{RobustNeRF} This dataset~\cite{Sabour_2023_CVPR} consists of natural and synthetic scenes designed to evaluate robustness at varying levels of visual distraction. The natural portion includes seven real-world scenes captured across streets, an apartment, and a robotics lab, where distractor objects are deliberately moved or allowed to move between frames to simulate long-term or uncontrolled capture conditions. These scenes vary widely in complexity, containing anywhere from a single distractor to up to 150 distractors, and include cases with strong view-dependent effects such as \texttt{Street1}, \texttt{Street2}, and \texttt{Gloss}. Frames are captured without temporal ordering and additional distractor-free images are provided for quantitative evaluation. To complement these real scenes, the dataset also includes synthetic sequences generated with the Kubric engine, where simple geometric objects are placed in a texture-less room, and a subset of them move between frames. By controlling object count, size, and motion, the synthetic setups allow precise analysis of distraction levels and their effect on RobustNeRF's performance.

\subsection{Training Strategies}
The proposed method is implemented in PyTorch, based on the 3D Gaussian Splatting framework gsplat~\cite{ye2025gsplat}.
The Gaussian primitives are initialized using approximately 500,000 point-cloud points predicted by VGGT-X, camera parameters are also taken from VGGT-X, and the scene scale is defined as the maximum distance from any camera position to the mean camera location. Each Gaussian has learnable parameters: mean, log-scale, quaternion rotation, opacity logit, and degree-3 spherical harmonic (SH) color coefficients. The learning rates (LR) for these parameters are summarized in Table~\ref{tab:lr}. An exponential LR scheduler decays the mean LR to 1\% over 7,000 steps, with evaluations at 4,000 and 7,000 steps, and SH degree increases by 1 every 1,000 steps. Most of the hyperparameters are chosen from 3DGS~\cite{kerbl3Dgaussians}, 3RGS~\cite{huang20253rgsbestpracticeoptimizing} and VGGT-X~\cite{liu2025vggtxvggtmeetsdense}. Whereas, a few like the number of total steps and the weights of various losses are manually adjusted to get the best results.

The MLP-based module uses an LR of 1.5e-4. For the appearance-embedding module used in MLP-based pose optimization, the MLP width is 32 with a depth of 2, with an LR of 1e-2 for the embedding and 1e-3 for the head. $\lambda_{\text{SSIM}}$ is the weight for the SSIM loss and is set to ${0.2}$, and $\lambda_{\text{Depth}}$, the depth loss weight is set to ${0.05}$. Global alignment of VGGT poses provides a consistent coordinate system, while per-image local pose refinement during evaluation runs for 100 iterations with an LR of 8e-4, decaying exponentially to 1\%. This refinement uses a gradient-masked L1 objective.

The near and far planes are set relative to the scene scale (near = 0.05 × scene scale, far = 1e10). At each evaluation step, memory usage, Gaussian count, PSNR~\cite{hore2010image}, SSIM~\cite{1284395}, and LPIPS~\cite{zhang2018perceptual} are recorded. Also, the Absolute Trajectory Error (ATE), which measures the difference between an estimated trajectory and its ground truth by first aligning the two trajectories to minimize the error, and then computing the error at each point and averaging it, often as a Root Mean Square Error (RMSE), along with the rotational errors are measured for the camera poses. This training setup jointly optimizes geometry, radiance (through staged SH activation), and camera parameters, with controlled scheduling, lightweight per-image refinement, and optional correspondence-based constraints.

For calculating ATE and rotational error, COLMAP optimized camera parameters are used. The extrinsic from the COLMAP is transferred to the predicted coordinates using the Kabsch-Umeyama algorithm~\cite{kabsch1976solution, kabsch1978discussion, umeyama2002least}. This is a method for finding the optimal transformation (translation, rotation, and scaling) to align two sets of corresponding points by minimizing the root-mean-square deviation (RMSD). This translation, rotation and scale are used to scale the COLMAP optimized extrinsics and are used as ground-truth.

\subsection{Results}
All experiments were conducted on a PC with an NVIDIA RTX 4090 GPU and 64GB RAM. 
Table~\ref{tab:baseline_mcmc} summarizes the baseline training performance across diverse indoor and outdoor scenes from Mip-NeRF 360, Tanks and Temples, and RobustNeRF. All runs use 1M Gaussians, with 500,000 Gaussians initialized from VGGT-X. Memory differences stem from per-scene activation and view count. Higher PSNR/SSIM and lower LPIPS/ATE/RTE indicate better reconstruction and pose accuracy.

Table~\ref{tab:nvs} compares novel view synthesis performance against 3DGS~\cite{kerbl3Dgaussians}, ZeroGS~\cite{yuchen2024zerogs}, and 3RGS~\cite{huang20253rgsbestpracticeoptimizing} across four scenes from Mip-NeRF 360. The proposed method achieves competitive PSNR, SSIM, and LPIPS scores, demonstrating high-fidelity reconstructions with better perceptual quality (LPIPS) while only training for 7,000 iterations, taking an average time of 180 seconds. Notably, it outperforms other techniques in LPIPS score, highlighting the effectiveness of the SfM-free pipeline and joint optimization strategy. Table~\ref{tab:pose} compares the pose errors for the same scenes using the same methods. Some results falls short of baselines because VGGT-X is not as robust as COLMAP for finding the intrinsics and extrinsics, also we are only training for 7,000 steps instead of 30,000 steps as normally used by other splatting based methods. 

Fig.~\ref{fig:rendering} presents qualitative rendering results from the proposed method on various scenes from selected datasets. The renderings are from the validation dataset. The images demonstrate high-fidelity reconstructions with accurate geometry and appearance details, showcasing the effectiveness of the SfM-free 3DGS pipeline.

\subsection{Ablation Studies}
To quantify the contribution of each component, ablation studies were conducted based on: (1) 3DGS as an MCMC sampler and (2) depth-guided loss, which is summarized in Table~\ref{tab:our_ablation}. 
Evaluations are conducted on selected scenes from the datasets, reporting average metrics for both novel view synthesis and camera pose estimation. 
The results highlight substantial improvements from the incorporation of each component. 
Although the baseline 3DGS method already performs camera pose optimization during training, its performance remains limited without the proposed additions. 

\section{Conclusion}

This paper introduced a novel pipeline that integrates 3DFMs with 3DGS to achieve fast and high-fidelity 3D scene reconstruction without relying on traditional SfM. Using VGGT-X for camera pose and depth estimation, the proposed method effectively initializes the Gaussian Splatting process, reducing optimization time to approximately 180 seconds on average while achieving a validation PSNR of 23.61 dB and LPIPS of 0.19. It is also found that in case the images are from high field of view camera with high barrel distortion, the VGGT-X model does not appropriately infer the camera parameters, thereby resulting in poor reconstruction.

The proposed MLP-based pose refinement and depth-guided loss further enhance reconstruction quality, particularly in low-viewpoint scenarios. Ablation studies confirm that both MCMC-based optimization and depth guidance contribute significantly to performance. Extensive experiments on Mip-NeRF 360, Tanks and Temples, and RobustNeRF demonstrated competitive novel view synthesis and camera pose accuracy (ATE of 0.016 m) compared to existing methods.

Future work will explore extending this framework to dynamic scenes and integrating semantic understanding for enriched 3D and 4D reconstructions, and improving on foundation model framework that can take different types of camera models.
%%%%%%%%%%%%%%%%%%%%%%%%%%%%%%%%%%%%%%%%%%%%%%%%%%%%%%%%%%%%%%%%%%%%%%%%%%%%%%%%
\bibliographystyle{ieeetr}
\bibliography{references}

\end{document}